\documentclass[10pt,twocolumn,letterpaper]{article}

\usepackage[pagenumbers]{cvpr} 

\definecolor{cvprblue}{rgb}{0.21,0.49,0.74}
\usepackage[pagebackref,breaklinks,colorlinks,allcolors=cvprblue]{hyperref}
\usepackage[accsupp]{axessibility}

\usepackage{url}            
\usepackage{booktabs}       
\usepackage{amsfonts}       
\usepackage{nicefrac}       
\usepackage{microtype}      
\usepackage{xcolor}         
\usepackage{graphicx,wrapfig,amsmath}
\usepackage[noend]{algorithm2e}
\usepackage[noend]{algpseudocode}
\usepackage{soul}

\usepackage{pifont}
\usepackage{xspace}
\makeatletter
\DeclareRobustCommand\onedot{\futurelet\@let@token\@onedot}
\def\@onedot{\ifx\@let@token.\else.\null\fi\xspace}
\def\eg{\emph{e.g}\onedot} 
\def\ie{\emph{i.e}\onedot}

\def\wrt{w.r.t\onedot} 

\usepackage{makecell}

\usepackage{amssymb}
\usepackage{color,colortbl}

\usepackage{multirow}
\usepackage{pifont}
\usepackage{gensymb}
\usepackage{enumitem}

\definecolor{lightgreen}{RGB}{100,220,100}
\definecolor{darkgreen}{RGB}{30,150,30}
\definecolor{darkblue}{RGB}{0,0,127}
\definecolor{darkyellow}{RGB}{171,133,0}
\definecolor{darkred}{RGB}{180,20,20}
\definecolor{darkmagenta}{RGB}{127,0,127}
\definecolor{darkcyan}{RGB}{0,127,127}

\definecolor{purple}{HTML}{9900ff}
\definecolor{darkpink}{HTML}{ff00ff}
\definecolor{maroon}{HTML}{980000}

\definecolor{lightred}{RGB}{220,20,20}

\newcommand{\darkgreen}[1]{\textcolor{darkgreen}{#1}}

\usepackage{stmaryrd}

\newcommand{\cvprcolor}[1]{\textcolor{cvprblue}{#1}}

\newcommand{\revision}[1]{#1}

\usepackage{lipsum}
\usepackage{bbm}

\newcommand{\expectation}{\mathop{\mathbb{E}}}

\definecolor{darkpastelgreen}{rgb}{0.01, 0.75, 0.24}

\newcommand{\sae}{\text{sae}}
\newcommand{\cb}{\text{cb}}

\title{Interpretable and Steerable Concept Bottleneck Sparse Autoencoders}

\author{
Akshay Kulkarni$^{1}$ \quad Tsui-Wei Weng$^{1}$ \quad Vivek Narayanaswamy$^{2}$\\
Shusen Liu$^{2}$ \quad Wesam A. Sakla$^{2}$ \quad Kowshik Thopalli$^{2}$\\[4pt]
$^{1}$University of California, San Diego \quad $^{2}$Lawrence Livermore National Laboratory\\[2pt]
{\tt\small \{a2kulkarni,lweng\}@ucsd.edu \quad \{narayanaswam1,liu42,sakla1,thopalli1\}@llnl.gov}
}
\begin{document}
\maketitle
\begin{abstract}
Sparse autoencoders (SAEs) promise a unified approach for mechanistic interpretability, concept discovery, and model steering in LLMs and LVLMs. However, realizing this potential requires learned features to be both interpretable and steerable. To that end, we introduce two new computationally inexpensive interpretability and steerability metrics for a systematic analysis of LVLM SAEs. This uncovers two observations; (i) a majority of SAE neurons exhibit either low interpretability or low steerability or both, rendering them ineffective for downstream use; and (ii) 
user-desired concepts are often absent in the SAE, thus limiting their practical utility. To address these limitations, we propose Concept Bottleneck Sparse Autoencoders (CB-SAE)\footnote{\revision{Code: \href{https://github.com/Trustworthy-ML-Lab/CB-SAE}{github.com/Trustworthy-ML-Lab/CB-SAE}}}—a novel post-hoc framework that prunes low-utility neurons and augments the latent space with a lightweight concept bottleneck aligned to a user-defined concept set. The resulting CB-SAE improves interpretability by +32.1\% and steerability by +14.5\% across LVLMs and image generation tasks. 
\end{abstract}
  
\vspace{-4mm}
\section{Introduction}
\label{sec:intro}

Sparse autoencoders (SAEs)~\cite{karvonen2025saebench, rajamanoharan2024jumping, saelens2024} have emerged as a foundational tool for mechanistic interpretability, mapping dense polysemantic activations into sparse monosemantic latents in large language models (LLMs)~\cite{lieberum2024gemma}, vision models\revision{~\cite{stevens2025towards,stevens2025saevision, fel2025archetypal, thasarathan2025universal}}, and large vision-language models (LVLMs)~\cite{pach2025sparse}.
However, realizing this promise requires SAE features to be semantically meaningful and causally effective, \ie \textit{interpretable} and \textit{steerable} respectively. 

\begin{figure}
    \centering
    \includegraphics[width=\columnwidth]{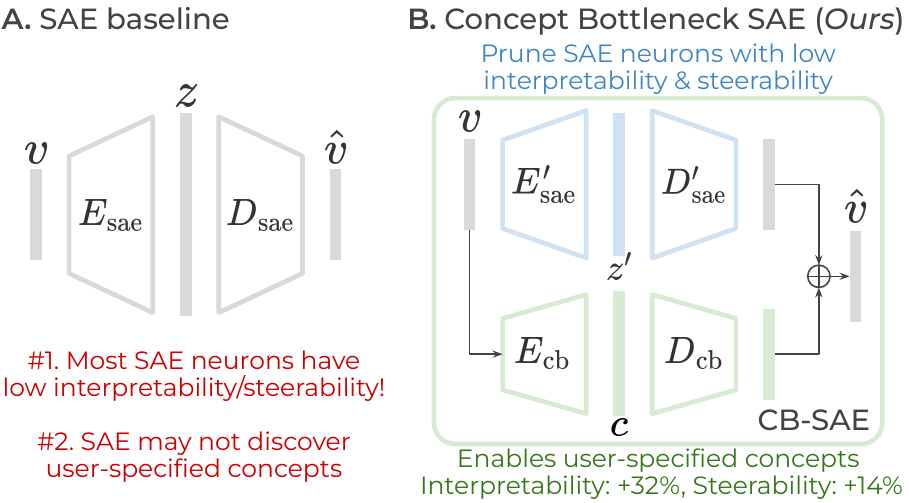}
    \vspace{-5mm}
    \caption{\textbf{A.} We find that the majority of SAE neurons in vision models have low interpretability or steerability, with no guarantee of discovering user-specified concepts. \textbf{B.} Our CB-SAE addresses both limitations by pruning SAE neurons with low interpretability and steerability, and replacing them with a user-specified concept bottleneck that improves both interpretability and steerability.}
    \vspace{-3mm}
    \label{fig:teaser}
\end{figure}

Recent work on SAEs~\cite{arad2025saes,wang2025does} in the context of LLMs shows that interpretability does not guarantee steering effectiveness, \ie features that activate strongly for a human-understandable concept may fail to control it when intervened upon~\revision{\cite{arad2025saes,wu2025axbench,karvonen2025saebench, kantamneni2025are}}. Although this trade-off has been observed in language models, its presence and implications in vision encoders and LVLMs remain largely unexplored. To investigate this in the LVLM setting, we conducted an empirical study of SAEs trained on activations from the CLIP~\cite{radford2021learning} vision encoder. We introduced metrics to quantify both interpretability and steerability at the neuron level and analyze their alignment across the SAE’s latent space. Our findings reveal that only about 19\% of SAE neurons exhibit both high interpretability and steerability.  Moreover, despite the SAE’s large dictionary size (65,536 neurons), it fails to represent 27–45\% of concepts drawn from established ImageNet-derived benchmarks~\cite{subramanyam2024decider}, even when trained on the corresponding data. This highlights the inability of unsupervised SAEs to cover user-specific concepts reliably. 

These findings surface two key limitations that constrain the practical utility of SAEs: (i) the inability to ensure comprehensive coverage of semantically meaningful concepts, and (ii) the lack of mechanisms for explicitly encoding user-defined concepts into the latent space to support better steerability. As a result, practitioners are left to work with the latent features the SAE happens to discover and searching post hoc for relevant activations, with no guarantee of alignment with task-specific requirements. This motivates the need for a unified framework that supports both unsupervised discovery and user-guided specification. 

\begin{figure}
    \centering
    \includegraphics[width=\columnwidth]{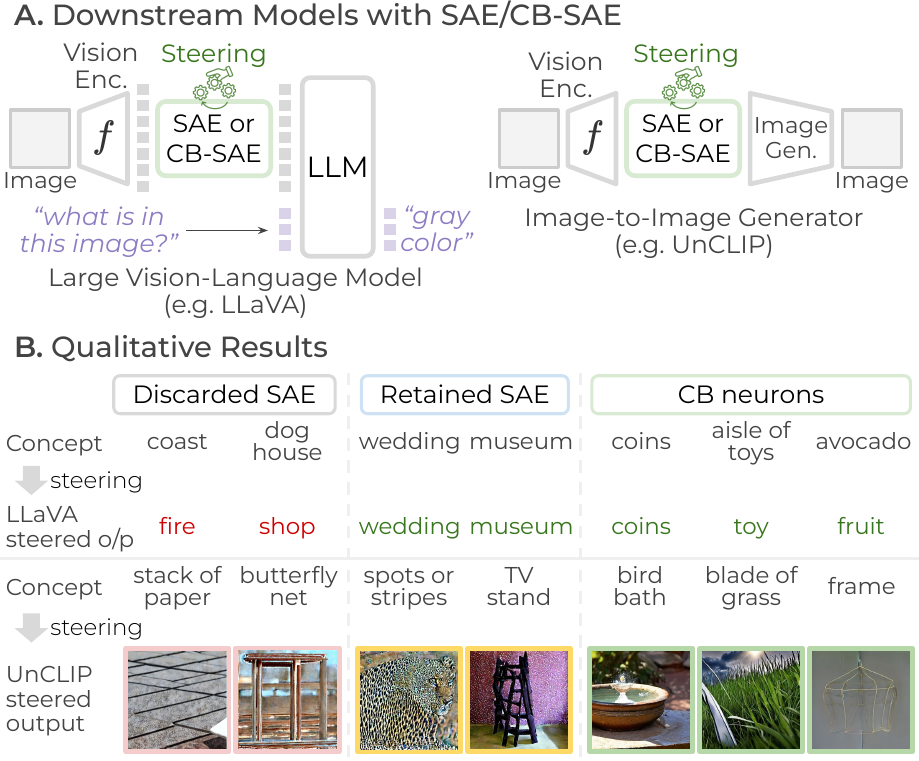}
    \vspace{-7mm}
    \caption{\textbf{A.} Our CB-SAE and baseline SAE can steer multiple downstream models like large vision-language models (LLaVA \cite{liu2024improved}) or image generative models (UnCLIP \cite{Rombach_2022_CVPR}). \textbf{B.} Examples of steering LLaVA and UnCLIP when using unit vector steering (zeroing out all SAE/CB-SAE neurons except the selected concept).}
    \label{fig:teaser_part2}
    \vspace{-3mm}
\end{figure}

A counterpart to SAEs are Concept Bottleneck Models (CBMs)~\cite{koh2020concept,yuksekgonul2023posthoc,oikarinen2023labelfree,srivastava2024vlg,sun2025concept,kulkarni2025interpretable}, with concept learning from a supervised perspective. CBMs explicitly train a model to predict a fixed set of human-interpretable concepts by introducing a bottleneck layer that mediates the final prediction. This enables guaranteed concept coverage, but limits CBMs to predefined concepts, preventing discovery of novel features unlike SAEs. These complementary strengths highlight the need for a unified framework that combines CBMs’ controllability with SAEs’ discovery capabilities.

Motivated by these observations, we propose Concept Bottleneck Sparse Autoencoders (CB-SAE) -- a unified framework that combines the unsupervised discovery capabilities of SAEs with the controllability of concept bottlenecks. We begin by pruning SAE features that lack interpretability and steerability, and then augment the resulting latent space with a lightweight CB autoencoder~\cite{kulkarni2025interpretable}, trained to align with a user-specified concept set (Fig.\ \ref{fig:teaser}\cvprcolor{B}). The model is optimized using tailored loss functions that preserve reconstruction fidelity, interpretability, and steerability. As a result, our CB-SAE produces latent features that are both semantically meaningful and causally effective.

We evaluate CB-SAE on two challenging downstream tasks: controlled text generation via vision–language models (LLaVA-1.5-7B~\cite{liu2023visual}, LLaVA-MORE~\cite{cocchi2025llava}) and controlled image synthesis using UnCLIP~\cite{Rombach_2022_CVPR}. CB-SAE consistently outperforms standard SAEs, with average gains of +32.1\% in interpretability and +14.5\% in steerability across all models and metrics. This improved performance is also shown qualitatively in Fig.\ \ref{fig:teaser_part2}\cvprcolor{B}, where the retained SAE and CB neurons consistently outperform discarded SAE neurons \wrt steerability.
To our knowledge, CB-SAE is the first framework to unify sparse autoencoders with concept bottleneck models, enabling robust interpretation and control of vision representations across modalities and architectures.

\section{Related Work}
\label{sec:related_work}




\noindent
\textbf{Sparse Autoencoders.}
SAEs 
aim to discover
interpretable features in neural networks by learning overcomplete decompositions of activations~\cite{makhzani2013ksae}. 
Recent work \cite{bricken2023monosemanticity, huben2024sparse} showed 
SAEs can decompose LLM representations into monosemantic features.
Various architectural innovations improved SAEs, like Batch-Top-$k$ sparsity~\cite{bussmann2024batchtopk}, JumpReLU~\cite{rajamanoharan2024jumping}, and Matryoshka SAEs \cite{bussmann2025learning} with multi-level feature hierarchies \revision{\cite{costa2025from}}. 
Large-scale efforts trained LLM SAEs across multiple layers and models~\cite{lieberum2024gemma, gao2025scaling}, with 
systematic benchmarks \cite{karvonen2025saebench}.
However, we uncover two key limitations of SAEs: 
they do
not guarantee the discovery of user-desired concepts, and many SAE neurons exhibit low interpretability or utility in downstream steering \revision{\cite{arad2025saes,fel2025archetypal,wu2025axbench,karvonen2025saebench, kantamneni2025are,yan2025visual}}. 

\noindent
\textbf{Concept Bottleneck Models.}
CBMs \cite{koh2020concept,yuksekgonul2023posthoc} provide a framework for building interpretable models by constraining predictions through a human-understandable concept layer, enabling both interpretation and steering. This approach has been extended to label-free settings~\cite{oikarinen2023labelfree}, enhanced with vision-language guidance~\cite{yan2023learning,yang2023language,srivastava2024vlg}, applied to image generative models \cite{ismail2024concept,kulkarni2025interpretable} as well as LLMs \cite{sun2025concept}.
\revision{DN-CBM \cite{rao2024discover} used an SAE encoder as a concept bottleneck layer, and trained a linear layer to classify using the SAE concepts. However, their approach is limited to image classification and susceptible to SAE limitations highlighted in our work.}
A concurrent work, AlignSAE \cite{yang2025alignsae}, independently devised a similar approach to introduce supervised concepts in SAEs. They attempt to disentangle the supervised concepts from the unsupervised SAE neurons with an orthogonality loss, while our approach explicitly prunes the low utility SAE neurons and only introduces the supervised concepts absent from the retained SAE neurons. 
Further, AlignSAE focuses on text-based LLM SAEs while we focus on vision SAEs for multimodal LLMs and image-to-image generative models.
Our work bridges SAEs and CBMs into our novel CB-SAE, combining the expressiveness of overcomplete feature decomposition with user-specified concepts, steerability, and interpretability of concept-guided learning.

\noindent
\textbf{SAEs for Vision and Vision-Language Models.} 
Recent work showed that SAEs can learn interpretable, monosemantic features in vision models \cite{stevens2025saevision} as well as vision-language models~\cite{pach2025sparse, zaigrajew2025interpreting}. 
\revision{SpLiCE \cite{bhalla2024interpreting} interprets CLIP with sparse concept embeddings using supervised reconstruction and a sparsity constraint, but does not support unsupervised concept discovery unlike SAE/CB-SAE.}
Another line of work \cite{venhoff2025visual, neotowards, papadimitriou2025interpreting} investigated how visual information maps to language via SAEs for cross-modal interpretability \cite{nasiri2025sparc,lou2025saevlm}. However, prior works neither address the challenges of ensuring discovered features are both interpretable and steerable, nor guarantee the discovery of user-specified concepts. Our CB-SAE addresses both limitations through post-hoc pruning and concept bottleneck training.

\section{Background}
\label{sec:back}


\begin{figure*}
    \centering
    \includegraphics[width=0.9\linewidth]{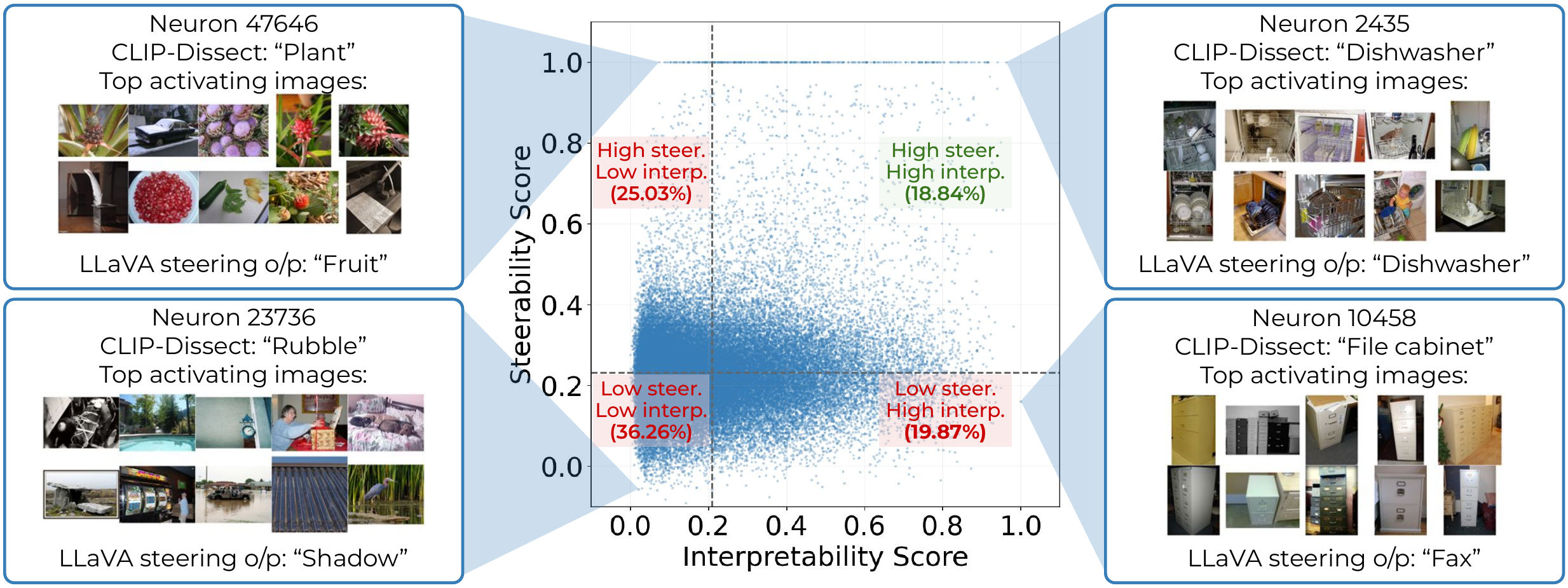}
    \vspace{-3mm}
    \caption{We analyze the interpretability and steerability of 65,536 neurons of an SAE trained for a CLIP image encoder. We also visualize the CLIP-Dissect assigned concept, top-activating images, and LLaVA steering outputs for some characteristic neurons. The dashed lines indicate the average scores along each axis, and we observe that most SAE neurons have either low interpretability, low steerability, or both.}
    \label{fig:analysis}
    \vspace{-3mm}
\end{figure*}

\textbf{SAE preliminaries.} 
Let $v = f_l(x) \in \mathbb{R}^d$ denote the dense activations from layer $l$ of a deep pre-trained vision model (e.g., CLIP image encoder~\cite{radford2021learning}) $f$ for an input image $x \in \mathbb{X}$. Here $d$ denotes the activation dimension and $\mathbb{X}$ corresponds to the space of images. SAEs decomposes the polysemantic activations $v$ into sparse, overcomplete latent representations $z \in \mathbb{R}^\omega$ ($\frac{\omega}{d} \gg 1$) with the aim of associating every unit in $z$ to distinct, interpretable concepts. Here $\frac{\omega}{d}$ corresponds to the expansion factor of the SAE~\cite{gao2025scaling}. Formally, an SAE is parameterized by a linear encoder $E_\sae \in \mathbb{R}^{\omega \times d}$, a linear decoder $D_\sae \in \mathbb{R}^{d \times \omega}$ , a shared bias term $b \in \mathbb{R}^d$, and a non-linear activation function $\sigma_\sae: \mathbb{R}^\omega \to \mathbb{R}^\omega$:
\begin{align}
    z &= \sigma_\sae(E_\sae (v-b)) \label{eqn:sae_enc} \\
    \hat{v} &= D_\sae z + b \label{eqn:sae_dec}
\end{align}
The SAE training objective is given by $\mathcal{L}_r = ||v - \hat{v}||_2^2 + \lambda||z||_1$, where $\lambda \geq 0$ balances reconstruction fidelity and sparsity, where $\hat{v}$ represents the SAE reconstruction. In addition to standard $\ell_1$ regularization, sparsity can be enforced directly via the activation function  $\sigma_\sae(\cdot)$, such as top-$k$~\cite{gao2025scaling}, batch top-$k$~\cite{bussmann2024batchtopk}, or ReLU with a learnable threshold~\cite{rajamanoharan2024jumping, lieberum2024gemma}.

\vspace{1mm}
\noindent
\textbf{Measuring SAE interpretability.} After training, SAEs are typically evaluated using reconstruction fidelity or sparsity~\cite{karvonen2025saebench, lou2025saevlm}. However, these metrics do not directly quantify interpretability which is namely the extent to which individual SAE neurons correspond to human-understandable concepts. While existing work~\cite{pach2025sparse, kim2025revelio} relies on manual inspection of top-$k$ activating inputs or autointerpretability scores~\cite{bills2023language, paulo2025automatically} that depend on external language models or measuring the monosemanticity~\cite{pach2025sparse}, they are often subjective, computationally expensive, and difficult to scale. 
To this end, we leverage a popular neuron interpretability tool CLIP-Dissect~\cite{oikarinen2023clip}, which utilizes a user-specified concept set $\mathcal{C}$ and a pretrained vision-language model to assign each neuron $j$ of an SAE to a human-interpretable text concept $c_j$. 
\revision{Our interpretability score is the similarity score computed in CLIP-Dissect.
This approach is computationally inexpensive, scalable, and is used for the first time with SAEs.}
\revision{Neuron interpretability scores have also been useful for downstream tasks like adversarial defense \cite{kulkarni2024igdefense}.}
Please refer to Appendix for further details on CLIP-Dissect.  



\vspace{1mm}
\noindent
\textbf{Measuring SAE steerability.} Beyond interpretability, SAEs have been shown to enable controllable manipulation of model behavior across language~\cite{arad2025saes}, large-scale vision~\cite{stevens2025saevision}, and large vision-language models~\cite{pach2025sparse,lou2025saevlm} such as LLaVA~\cite{liu2024improved}. Steerability refers to the ability to influence model outputs through targeted modifications of SAE neuron activations, thereby inducing semantically consistent changes~\cite{arad2025saes}. It is typically quantified by measuring the alignment between the steered output and the concept label of the intervened neuron. Since, in our study, the base model $f$ is a vision-only encoder, we employ a downstream generative model to evaluate the effect of SAE latent interventions.

Following~\cite{pach2025sparse}, we adopt LLaVA \cite{liu2024improved}, which maps an image–text pair $(x, t)$ to a text output $o$. In LLaVA, the vision encoder $f$ (\eg CLIP \cite{radford2021learning}) produces visual tokens $\{v_i\}_{i=1}^{N}$, which are projected by an adapter into the LLM’s word embedding space, where they are combined with prompt tokens to generate $o$. Similar to \cite{pach2025sparse}, to probe steerability, we use a white image with the prompt \textit{``What is shown in this image? Use exactly one word!''}. For a given target neuron $j \!\in\! [\omega]$ of the SAE, we overwrite its activation across all tokens (from this white image) with a fixed value $\alpha$, reconstruct the modified latents $\{\tilde{v}_i\}$ through the SAE decoder, and feed them into LLaVA to produce the steered output $\tilde{o}_j$.

We then compute the cosine similarity between $\tilde{o}_j$ and the neuron's CLIP-Dissect-assigned concept $c_j$ in a sentence-transformer embedding space~\cite{reimers-2019-sentence-bert}. Higher similarity indicates greater steerability, as the neuron reliably drives the output toward its associated concept. Unlike~\cite{pach2025sparse}, which compared steered outputs to top-activating images in CLIP’s image-text space, our method compares $o_j$ with concepts identified by CLIP-Dissect, which produces these descriptions by aggregating activations across all images thus yielding a more robust, semantically grounded steerability metric. 
Note that while steering (image,text)-to-text LLaVA \cite{liu2024improved} is one way to compute a steerability metric, a similar metric can be computed using an image-to-image generator like UnCLIP \cite{Rombach_2022_CVPR} (see Appendix for analysis with UnCLIP).

\section{Interpretability vs Steerability in SAEs}
\label{sec:analysis_expt}
In this section, we empirically analyze SAEs from two complementary perspectives: (i) their capacity to capture interpretable concepts and (ii) their ability to steer model outputs and quantify their trade-offs. While these two measures \ie, interpretability and steerability are related, they capture distinct yet complementary aspects of SAE behavior~\cite{wang2025does}. 
In practice, interpretable neurons may not be steerable if their activations are weakly causal or entangled, while steerable neurons may encode abstract features misaligned with user objectives~\cite{arad2025saes,wang2025does}. The insights from this analysis on large vision-language models motivate our hybrid framework, introduced in Sec.\ \ref{sec:approach}, which integrates SAEs with principles from concept bottleneck models~\cite{koh2020concept,oikarinen2023labelfree,srivastava2024vlg}.

\noindent\textbf{Expt.\ 1: Are all SAE neurons interpretable \& steerable?}

\noindent\underline{Setup}. We train a Matryoshka Batch Top-$k$ SAE~\cite{bussmann2024batchtopk,zaigrajew2025interpreting} with $\omega$ = 65536 and an expansion factor of 64 on layer $l$ = 22 of the CLIP-ViT-L/14-336 vision encoder~\cite{radford2021learning}, following the setup of~\cite{pach2025sparse}, using the ImageNet-1K dataset~\cite{deng2009imagenet} (see Appendix for more details and results with other layers and models). As CLIP-Dissect requires a predefined concept set $\mathcal{C}$, we employ the Broden dataset~\cite{netdissect2017}, ($|\mathcal{C}|$=1197) which provides both low-level attributes and object-level visual concepts. We then compute interpretability and steerability scores for each SAE neuron as described in Sec.\ \ref{sec:back}. 

\begin{figure*}
    \centering
    \includegraphics[width=\linewidth]{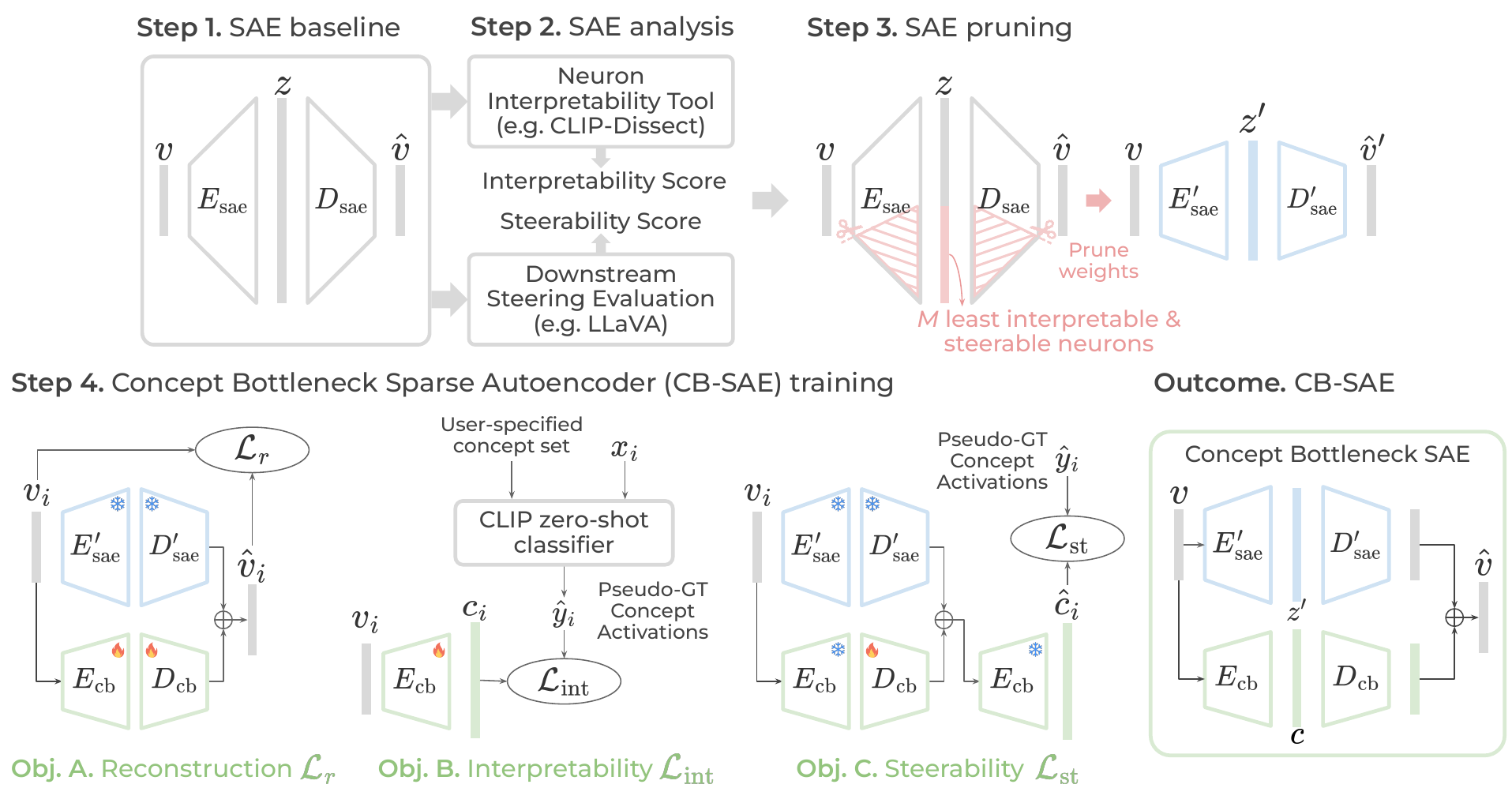}
    \vspace{-5mm}
    \caption{\textbf{Pipeline for building CB-SAE}. \textbf{Step 1.} A baseline SAE is trained and \textbf{Step 2.} evaluated with CLIP-Dissect and downstream steering to obtain interpretability and steerability scores per SAE neuron. \textbf{Step 3.} The $M$ least interpretable and steerable neurons are pruned by deleting the corresponding SAE weights. \textbf{Step 4.} We train the CB-SAE with frozen, pruned SAE weights with three objectives: \textbf{A.} recover the reconstruction ability lost by pruning using $\mathcal{L}_r$, \textbf{B.} incorporate the user-specified concept set with $\mathcal{L}_\text{int}$, and \textbf{C.} promote steerability with a cyclic reconstruction loss $\mathcal{L}_\text{st}$. \textbf{Outcome.} CB-SAE uses both retained SAE and CB-AE for inference.}
    \label{fig:approach}
    \vspace{-2mm}
\end{figure*}

\noindent\underline{Observations}. Fig.~\ref{fig:analysis} illustrates the trade-off between interpretability and steerability scores, with dashed horizontal and vertical lines marking their respective mean values to delineate four distinct neuron groups. For each group, the figure shows the top-10 activating images, CLIP-Dissect-assigned concepts (interpretability), and LLaVA-steered outputs (steerability) for a representative neuron, highlighting characteristic behaviors. The distribution of neurons across these groups is as follows:

\begin{itemize}
\item \textit{Low interpretability, low steerability} (36.26\%, 23,763): These neurons are largely inactive and contribute minimally to either semantic meaning or controllable behavior.
\item \textit{High interpretability, low steerability} (19.87\%, 13,022): These neurons capture clear, human-understandable concepts but have limited influence on model outputs.
\item \textit{Low interpretability, high steerability} (25.03\%, 16,403): These neurons effectively steer outputs but correspond to abstract or composite features that lack semantic clarity.
\item \textit{High interpretability, high steerability} (18.84\%, 12,348): This is the most desirable group of neurons that are both interpretable and causally effective.
\end{itemize}
These indicate that a vast majority of SAE neurons are not directly useful for downstream tasks such as explanation or control, reinforcing the need for hybrid approaches that jointly enhance interpretability and steerability.

\noindent\textbf{Expt.\ 2: Can SAEs represent all user-specified concepts?} 
A key question is whether the SAE can represent all concepts from a given concept set, thereby supporting both human interpretability and controllable model behavior. Although the SAE contains 65,536 neurons, far exceeding the size of standard concept sets, its ability to represent concepts varies considerably with the diversity and complexity of the set. 

\noindent\underline{Observations}. 
Using CLIP-Dissect, we evaluate the coverage of unique concepts across multiple concept sets:
\begin{itemize}
\item Broden~\cite{netdissect2017}: 1,153/1,197 (96.3\%)
\item VLG-CBM~\cite{srivastava2024vlg}: 3,445/4,729 (72.8\%)
\item DECIDER~\cite{subramanyam2024decider}: 4,333/7,827 (55.3\%)
\item 3k common English words~\cite{oikarinen2023labelfree}: 1,857/3,000 (61.9\%)
\item 20k common English words~\cite{oikarinen2023labelfree}: 5,596/20,000 (28.0\%)
\end{itemize}

The SAE performs well on the smaller and well-defined Broden concept set, capturing 96.3\% of its visual concepts. However, coverage drops sharply for larger or linguistically diverse sets, with only 28–73\% of concepts represented on average. Notably, despite being trained on ImageNet, the SAE fails to capture 27–45\% of ImageNet-related concepts from the VLG-CBM and DECIDER sets. These results indicate that, while SAEs effectively capture simple, low-level concepts, their latent spaces struggle to generalize to broader, user-specified concept sets—limiting their utility for downstream interpretability and nuanced steerability.

Our analysis reveals two key requirements: (i) expanding the SAE latent space to capture a broader range of semantically distinct concepts while remaining effective for downstream tasks such as steering, and (ii) enabling explicit user specification of concepts within the SAE. Simply pruning neurons with low interpretability and steerability degrades reconstruction fidelity. 
To address this and enable user-specified concepts, we propose to train a \textit{concept bottleneck autoencoder}~\cite{koh2020concept,kulkarni2025interpretable} alongside the retained SAE. This hybrid framework combines (supervised) concept alignment with (unsupervised) discovery, restoring reconstruction fidelity while enhancing concept coverage and steerability.

\section{Our Approach: CB-SAE}
\label{sec:approach}


We propose a novel concept bottleneck sparse autoencoder (CB-SAE) based on our analysis to address two limitations of sparse autoencoders namely low interpretability/steerability and the lack of support for user-specified concepts.

\subsection{Pruning SAE neurons}

\noindent
\textbf{\underline{Step 1} (Fig.\ \ref{fig:approach})}
As discussed in Sec.\ \ref{sec:back}, we begin with training an SAE on layer $l$ activations from the vision model $f$.

\noindent
\textbf{\underline{Step 2} (Fig.\ \ref{fig:approach})}
Following our analysis experiments, we compute the interpretability and steerability scores for each sparse neuron in the trained SAE denoted by $I \in [0, 1]^\omega$ and $S\in [0, 1]^\omega$ respectively. 

\noindent
\textbf{\underline{Step 3} (Fig.\ \ref{fig:approach})}
We prune the SAE weights $E_\sae, D_\sae$ to remove the $M$ least interpretable and steerable SAE neurons as they are unsuitable for downstream applications. Concretely, the set of $M$ SAE neurons to be pruned is $\mathcal{P}=\{m \; | \; I_m + S_m < \tau, \; m\in [\omega]\}$ where $\tau$ is the threshold that determines $|\mathcal{P}|=M$ and $[\omega]=\{1, 2,\cdots,\omega\}$. In practice, we simply sort the $I+S$ in descending order and select the bottom-$M$ neurons that
constitute $\mathcal{P}$.

SAE consists of $E_\sae \in \mathbb{R}^{\omega\times d}$ and $D_\sae\in\mathbb{R}^{d\times \omega}$. We can then prune the selected set of neurons $\mathcal{P}$ by deleting the corresponding rows and columns in $E_\sae$ and $D_\sae$ respectively. In other words, the retained SAE weights $E'_\sae$ and $D'_\sae$ have all rows and columns other than those in $\mathcal{P}$ respectively:
\begin{align}
    E'_\sae &= E_\sae[[\omega] \setminus \mathcal{P}, :] \\
    D'_\sae &= D_\sae[:, [\omega] \setminus \mathcal{P}]    
\end{align}
Here, $\setminus$ denotes the set minus operator. Note, the shared bias term $b\in\mathbb{R}^d$ does not change as it is independent of the number of SAE neurons $\omega$. The retained SAE consists of the retained encoder $E'_\sae\in\mathbb{R}^{(\omega-M)\times d}$, retained decoder $D'_\sae\in\mathbb{R}^{d\times (\omega-M)}$, and the bias $b\in\mathbb{R}^d$. Using Eq.\ \eqref{eqn:sae_enc}, \eqref{eqn:sae_dec} with $E'_\sae, D'_\sae$, the retained SAE latent changes to $z'\in\mathbb{R}^{(\omega-M)}$ and the reconstructed latent is $\hat{v}'\in\mathbb{R}^d$ (Fig.\ \ref{fig:approach}, Step \cvprcolor{3}). While the reconstructed $\hat{v}'$ has the same dimensions as $v$, the average reconstruction loss $\expectation_v[v-\hat{v}']$ will be higher than without pruning $\expectation_v[v-\hat{v}]$, due to loss of information to effectively reconstruct the activations. As discussed earlier, to recover this lost reconstruction ability and to incorporate user-specified concepts, we introduce a concept bottleneck.

\subsection{Training CB-SAE}
\label{sec:training_cbsae}

\noindent
\textbf{\underline{Step 4} (Fig.\ \ref{fig:approach})}
We introduce a concept bottleneck autoencoder \cite{kulkarni2025interpretable} alongside the retained SAE (Fig.\ \ref{fig:teaser}\cvprcolor{B}). Our CB-SAE consists of the retained SAE $E'_\sae, D'_\sae$, a linear concept encoder $E_\cb\in\mathbb{R}^{|\mathcal{C}|\times d}$, a linear concept decoder $D_\cb\in\mathbb{R}^{d\times |\mathcal{C}|}$, and a non-linear activation function $\sigma_\cb:\mathbb{R}^{|\mathcal{C}|}\to\mathbb{R}^{|\mathcal{C}|}$ similar to the SAE, where $\mathcal{C}$ is a pre-defined concept set. For an input $v\in\mathbb{R}^d$, the CB-SAE reconstructs $\hat{v}'\in\mathbb{R}^d$ as,
\begin{align}
    z' &= \sigma_\sae(E_\sae'(v-b)) \\
    c &= E_\cb (v-b) \\
    \hat{v}' &= D'_\sae z' + b + D_\cb \sigma_\cb(c) \label{eqn:cbsae_dec}
\end{align}
We use a top-$k$ function as $\sigma_\cb$ with $k \ll |\mathcal{C}|$ to ensure that sparsity constraints are similar to the original SAE. The bias term $b\in\mathbb{R}^d$ is shared with the retained SAE.

\noindent
\textbf{Concept Set Selection.}
Based on our motivation to support user-specified concepts, the concept set $\mathcal{C}$ can be specified by the user as a list of text-based concepts, similar to prior work~\cite{oikarinen2023labelfree,srivastava2024vlg}. However, as shown in Sec.\ \ref{sec:analysis_expt}, the SAE can already represent some concepts well and including them in the concept set $\mathcal{C}$ would be redundant. Hence, we only use concepts absent from the retained SAE in the CB-SAE concept set. Let $\mathcal{C}_\text{user}$ be the user-specified concept set, $\mathcal{C}_\text{rsae}\subset \mathcal{C}_\text{user}$ be the concepts present in the retained SAE (found using CLIP-Dissect before pruning the SAE, Fig.\ \ref{fig:approach} Step \cvprcolor{2}). Then our CB-SAE concept set is given by $\mathcal{C}=\mathcal{C}_\text{user} \setminus \mathcal{C}_\text{rsae}$.

\vspace{1mm}
\noindent
\textbf{Training Objectives.}
We propose three training objectives to guide the CB-SAE. First, the concept bottleneck should recover the reconstruction fidelity lost by pruning the SAE neurons. Second, the CB neurons should be interpretable \wrt the concept set $\mathcal{C}$. And third, the CB neurons should be steerable \wrt the concept set $\mathcal{C}$.
The neurons of the retained SAE already meet the reconstruction, interpretability, and steerability objectives for their discovered concepts $\mathcal{C}_\text{rsae}$, so we keep the retained SAE weights frozen.

\vspace{1mm}
\noindent
\textbf{Objective A: Reconstruction $\mathbf{\mathcal{L}_r}$} (Fig.\ \ref{fig:approach}, Step \cvprcolor{4A}).
Similar to the SAE, we optimize the mean-squared error $\mathcal{L}_r$ between the input latent $v\in\mathbb{R}^d$ and the CB-SAE reconstruction $\hat{v}'$,
\begin{align}
    \min_{\textcolor{darkgreen}{{E_\cb, D_\cb}}} [\mathcal{L}_r(v, \hat{v}')]
\end{align}
Instead of an $\ell_1$ regularizer for sparsity, we use a top-$k$ activation function $\sigma_\cb$ as mentioned earlier.

\vspace{1mm}
\noindent
\textbf{Objective B: Interpretability $\mathbf{\mathcal{L}_\text{int}}$} (Fig.\ \ref{fig:approach}, Step \cvprcolor{4B}).
To ensure that each CB neuron in $c$ activates for its corresponding concept in the concept set $\mathcal{C}$, we use a CLIP zero-shot classifier~\cite{radford2021learning} with $\mathcal{C}$ as the classes and obtain pseudo-ground-truth concept activations similar to prior work in CBMs~\cite{oikarinen2023labelfree, kulkarni2025interpretable}. This enables concept-label-free training of the CB-SAE, \ie does not require explicit concept labels and also supports any arbitrary user-specified concept sets.

The zero-shot classifier $\mathcal{M}:\mathbb{X}\times\mathbb{T}^{|\mathcal{C}|} \to \mathbb{R}^{|\mathcal{C}|}$  ($\mathbb{T}$ refers to text space) takes in an image $x\in\mathbb{X}$, list of concept names $\mathcal{C}$, and predicts concept logits $\hat{y}=\mathcal{M}(x, \mathcal{C})\in\mathbb{R}^{|\mathcal{C}|}$. We use a cosine-cubed similarity loss $\mathcal{L}_\text{int}$ following \cite{oikarinen2023labelfree} between the concept encoder's predictions $c=E_\cb(v)$ and $\hat{y}$,
\begin{align}
    \min_{\textcolor{darkgreen}{{E_\cb}}} [\mathcal{L}_\text{int}(c, \hat{y})]
\end{align}
Here, $v=f_l(x)$ is the vision encoder $f$ output at layer $l$ for the same image $x$ used with $\mathcal{M}$. Further, note that we use the sparsity constraint of a top-$k$ activation function $\sigma_\cb$ only for the decoder in Eq.\ \eqref{eqn:cbsae_dec}. This is because the concept encoder $E_\cb$ should be able to interpret all concepts in an image, but the concept decoder $D_c$ can only use the top-$k$ concepts for reconstruction. This is further ensured by updating only $E_\cb$ with the interpretability objective $\mathcal{L}_\text{int}$. We defer more details of the cosine-cubed similarity loss \cite{oikarinen2023labelfree} to the Appendix.

\begin{table*}[t]
\centering
\caption{\textbf{Interpretability and Steerability Evaluation with LLaVA and UnCLIP.} All four metrics are in 0-1 range (higher is better), CD indicates CLIP-Dissect score and MS indicates monosemanticity score.}
\vspace{-3mm}
\label{tab:llava_main_table}
\begin{tabular}{@{}lllcccc@{}}
\toprule
\multirow{2}{*}{\begin{tabular}[c]{@{}l@{}}Downstream\\ Task\end{tabular}}&\multirow{2}{*}{\begin{tabular}[c]{@{}l@{}}Steered\\ Model\end{tabular}} & \multirow{2}{*}{Method} & \multicolumn{2}{c}{Interpretability} & \multicolumn{2}{c}{Steerability} \\ \cmidrule(lr){4-5}\cmidrule(lr){6-7} 
 & &  & CD & MS & Unit-Vec & White Image \\ \midrule
\multirow{4}{*}{\renewcommand{\arraystretch}{0.7}\begin{tabular}[c]{@{}l@{}}Image + Text $\to$ \\ Text Generation\end{tabular}} & \multirow{2}{*}{\renewcommand{\arraystretch}{0.7}\begin{tabular}[c]{@{}l@{}}LLaVA-1.5-7B \cite{liu2024improved} \\ 
(CLIP-ViT-L + Vicuna-7B)\end{tabular}} & SAE \cite{pach2025sparse} & 0.154 & 0.517 & 0.198 & 0.203 \\
 &  & \cellcolor{gray!10}CB-SAE (\textit{Ours}) & \cellcolor{gray!10}\textbf{0.244} & \cellcolor{gray!10}\textbf{0.556} & \cellcolor{gray!10}\textbf{0.261} & \cellcolor{gray!10}\textbf{0.250} \\
\cmidrule(lr){2-7}
 & \multirow{2}{*}{\renewcommand{\arraystretch}{0.7}\begin{tabular}[c]{@{}l@{}}LLaVA-MORE \cite{cocchi2025llava}\\(DINOv2-L + Gemma2-9B)\end{tabular}} & SAE \cite{pach2025sparse} & 0.194 & 0.553 & 0.179 & 0.177 \\
 &  & \cellcolor{gray!10}CB-SAE (\textit{Ours}) & \cellcolor{gray!10}\textbf{0.291} & \cellcolor{gray!10}\textbf{0.598} & \cellcolor{gray!10}\textbf{0.192} & \cellcolor{gray!10}\textbf{0.189} \\ \midrule 
\multirow{2}{*}{\renewcommand{\arraystretch}{0.7}\begin{tabular}[c]{@{}l@{}}Image $\to$ Image\\ Generation\end{tabular}} & \multirow{2}{*}{\renewcommand{\arraystretch}{0.7}\begin{tabular}[c]{@{}l@{}}UnCLIP \cite{Rombach_2022_CVPR} \\ (CLIP-ViT-L + SD-2.1)\end{tabular}} & SAE \cite{pach2025sparse} & 0.058 & 0.540 & 0.642 & 0.654 \\
 &  & \cellcolor{gray!10}CB-SAE (\textit{Ours}) & \cellcolor{gray!10}\textbf{0.092} & \cellcolor{gray!10}\textbf{0.594} & \cellcolor{gray!10}\textbf{0.659} & \cellcolor{gray!10}\textbf{0.664} \\ \bottomrule 
\end{tabular}
\end{table*}

\vspace{1mm}
\noindent
\textbf{Objective C: Steerability $\mathbf{\mathcal{L}_\text{st}}$} (Fig.\ \ref{fig:approach}, Step \cvprcolor{4C}).
Prior work on CBMs for generative models \cite{kulkarni2025interpretable} 
leveraged the downstream image generation task to design explicit steerability objectives.
In contrast, we propose a simple task-agnostic cyclic reconstruction objective for steerability. With this, we show in Sec.\ \ref{sec:expts} that the same CB-SAE can steer two different downstream tasks: image-to-image generation and image-text-to-text generation.

Concretely, as shown in Fig.\ \ref{fig:approach}, Step \cvprcolor{4C}, we pass the reconstructed latent $\hat{v}'$ back through the concept encoder $E_\cb$ to produce cyclically reconstructed concepts $\hat{c}=E_\cb(\hat{v}'-b)$. Then, we use the same pseudo-ground-truth concept activations $\hat{y}$ computed for Objective \cvprcolor{B} and optimize the same loss as Objective \cvprcolor{B} with $\hat{c}$ (denoted by $\mathcal{L}_\text{st}$ for clarity),
\begin{align}
    \min_{\textcolor{darkgreen}{{D_\cb}}} [\mathcal{L}_\text{st}(\hat{c}, \hat{y})]
    \label{eqn:st_loss}
\end{align}
Only the concept decoder $D_\cb$ is updated with the steerability loss. This is because only the decoder $D_\cb$ is responsible to appropriately modify the latent $\hat{v}'$ when a concept in $c$ is modified for steering. On the other hand, the concept encoder $E_\cb$ should focus only on interpreting the input $v$ and updating it with $\mathcal{L}_\text{st}$ could hurt interpretability.

Instead of loss weighting hyperparameters, we train by alternately minimizing the objectives
via separate Adam optimizers which adaptively scale weight updates \cite{kundu2019bihmp}.

\section{Experiments}
\label{sec:expts}

\begin{table}[t]
\centering
\caption{Evaluating interpretability and steerability of discarded SAE neurons, retained SAE neurons, and CB neurons separately.}
\vspace{-3mm}
\label{tab:neuron_ablation}
\resizebox{\linewidth}{!}{
\begin{tabular}{@{}lccc@{}}
\toprule
\multirow{2}{*}{Set of Neurons} & Interpretability & \multicolumn{2}{c}{Steerability} \\ \cmidrule(lr){2-2} \cmidrule(lr){3-4}
 & CLIP-Dissect & Unit-Vec & White Image \\ \midrule
All SAE neurons & 0.154 & 0.198 & 0.203 \\
Discarded SAE neurons & 0.084 & 0.144 & 0.162 \\
Retained SAE neurons & 0.238 & \textbf{0.263} & \textbf{0.252} \\
CB neurons & \textbf{0.323} & 0.231 & 0.219 \\
All CB-SAE neurons & 0.244 & 0.261 & 0.250 \\ \bottomrule
\end{tabular}
}
\vspace{-3mm}
\end{table}

\begin{figure*}[t]
    \centering
    \includegraphics[width=0.9\linewidth]{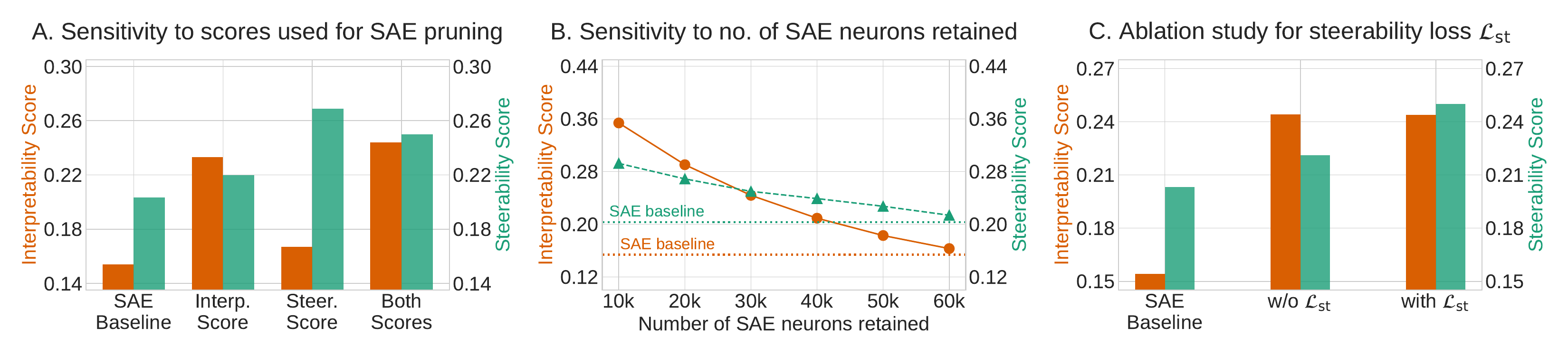}
     \vspace{-3mm}
    \caption{\textbf{A.} Sensitivity of CB-SAE to the choice of scores used for SAE pruning. \textbf{B.} Sensitivity of CB-SAE to the number of SAE neurons retained. \textbf{C.} Ablation study for our proposed steerability objective $\mathcal{L}_\text{st}$ from Eq.\ \eqref{eqn:st_loss}.}
    \vspace{-3mm}
    \label{fig:ablation_sens}
\end{figure*}


We extensively evaluate our proposed CB-SAE \wrt interpretability and steerability on two downstream tasks, (image,text)-to-text generation and image-to-image generation. We also performed detailed ablation and sensitivity analysis experiments to validate our design choices.

\subsection{Setup}

\noindent
\textbf{Baseline SAE and CB-SAE.} We follow \citet{pach2025sparse} and train a Matryoshka Batch Top-$k$ SAE \cite{zaigrajew2025interpreting} with expansion factor $\frac{\omega}{d}=64$ as the baseline SAE on the ImageNet-1k \cite{deng2009imagenet} dataset. Our CB-SAE is also trained on the same intermediate activations as the baseline SAE for a fair comparison.
We retain $\omega-M= 30$k neurons in the SAE pruning and use a top-$k$ function as $\sigma_\cb$ with $k=5$ in our CB-SAE. 
We use the VLG-CBM ImageNet concept set \cite{srivastava2024vlg,oikarinen2023labelfree} for the CB neurons.
In our training, we use a CLIP-ViT-B/16 \cite{radford2021learning} model for obtaining the pseudo-ground-truth concept activations.

\vspace{1mm}
\noindent
\textbf{Evaluation Metrics.} To evaluate interpretability, we use the CLIP-Dissect interpretability score introduced in Sec.\ \ref{sec:back} and the monosemanticity score from~\cite{pach2025sparse} using the ImageNet validation set. To ensure a fair evaluation, we use a stronger CLIP-ViT-L/14 model (\wrt smaller ViT-B/16 used for training CB-SAE). To evaluate steerability, we use our proposed steerability score (Sec.\ \ref{sec:back}). Concretely, we evaluate the steerability of each CB/SAE neuron in two ways:
\begin{itemize}
    \item Unit Vector: The selected neuron is activated to a high value $\alpha=50$ (as in \cite{pach2025sparse}) \& all other neurons are set to $0$.
    \item White Image: The selected neuron is activated to a high value $\alpha=50$ and all other neurons have the values predicted when using an empty white image (following \cite{pach2025sparse}) as input, instead of $0$ like in unit vector steering.
\end{itemize}
The interpretability and steerability scores of individual neurons are averaged to obtain the overall scores.
For experiments where the steered output is text, we compare the similarity between the steered text and the CLIP-Dissect assigned concept for the selected neuron in a sentence transformer embedding space (as in Sec.\ \ref{sec:back}). For experiments where steered output is an image, we compute the average similarity between the steered image and top-16 highly activating images for the selected neuron in the DINOv2~\cite{oquab2024dinov2} embedding space. This is because the diffusion model being steered (UnCLIP \cite{Rombach_2022_CVPR}) may rarely return partially or completely noisy images after steering, which cannot be properly evaluated with an image-text similarity score (\eg CLIP) that expects clean images. All metrics are normalized in 0-1 range and higher values indicate better performance.

\subsection{Quantitative Comparison}
In Table \ref{tab:llava_main_table}, we compare our CB-SAE with the baseline SAE \cite{pach2025sparse} across several downstream models as well as tasks. For two different variants of the (image,text)-to-text LLaVA \cite{liu2024improved,cocchi2025llava} model, our CB-SAE demonstrates consistent gains over the SAE baseline across both interpretability (avg.\ \darkgreen{+33.0\%} for CLIP+Vicuna and avg.\ \darkgreen{+29.0\%} for DINOv2+Gemma) and steerability metrics (avg.\ \darkgreen{+27.5\%} for CLIP+Vicuna and avg.\ \darkgreen{+{14.0}\%} for DINOv2+Gemma).
Similarly, for the image-to-image UnCLIP \cite{Rombach_2022_CVPR} generative model, our CB-SAE outperforms the SAE baseline (avg.\ \darkgreen{+34.3\%} interpretability and avg.\ \darkgreen{+{2.1}\%} steerability). To the best of our knowledge, we are the first to show that an SAE (and CB-SAE) trained with the same method can be used to steer different downstream tasks.

\begin{figure}[t]
    \centering
    \includegraphics[width=0.85\linewidth]{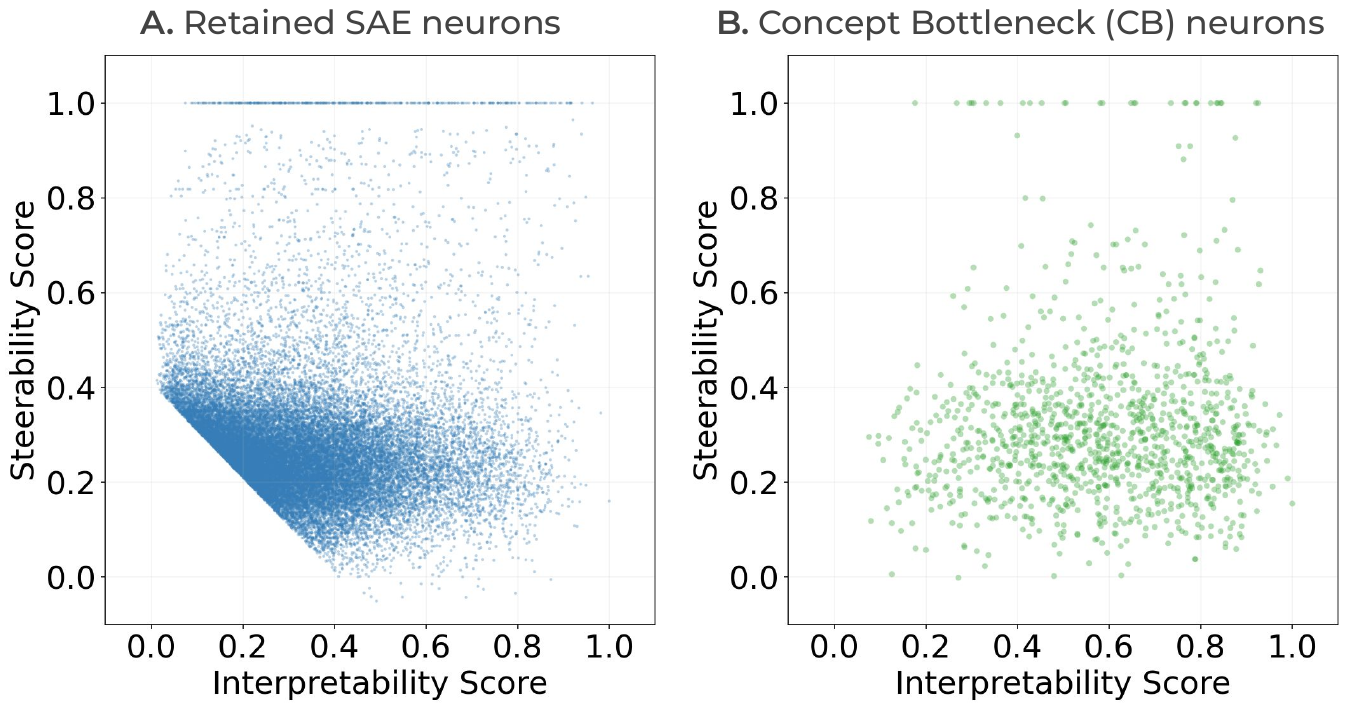}
    \vspace{-3mm}
    \caption{Visualizing the interpretability and steerability of retained SAE neurons and CB neurons, similar to Fig.\ \ref{fig:analysis}.}
    \vspace{-3mm}
    \label{fig:cbsae_scatter}
\end{figure}

\subsection{Analysis of our CB-SAE}

\textbf{Effect of CB neurons.}
In Table \ref{tab:neuron_ablation}, we separately evaluate the interpretability and steerability of discarded and retained SAE neurons, as well as CB neurons. We observe CB neurons have significantly higher interpretability than SAE neurons. Whereas, CB neurons' steerability is worse than retained SAE neurons but significantly better than discarded SAE neurons as well as all SAE neurons (discarded+retained). 
Intuitively, the retained SAE neurons contain many highly steerable neurons because steerability (and interpretability) were used to prune SAE neurons.

\noindent
\textbf{Sensitivity to scores used for SAE pruning.} In Fig.\ \ref{fig:ablation_sens}\cvprcolor{A}, we compare the CB-SAE performance while varying the choice of metrics for SAE pruning: either interpretability score or steerability score or both. We find that prioritizing either score leads to some loss in performance on the other, while using both scores gives balanced performance. This is beneficial as users can choose the score or even design a new score for pruning based on their target downstream usecase.

\noindent
\textbf{Sensitivity to no. of SAE neurons retained.}
In Fig.\ \ref{fig:ablation_sens}\cvprcolor{B}, we evaluate CB-SAE models trained with varying number of SAE neurons retained after pruning using both interpretability and steerability scores. We find that lower number of SAE neurons retained leads to higher interpretability and steerability scores. This is because pruning keeps a smaller subset of SAE neurons with higher scores. However, further reducing the number of SAE neurons would hurt performance as reconstruction becomes more difficult.

\noindent
\textbf{Ablation study for steerability loss $\mathbf{\mathcal{L}_\text{st}}$.} In Fig.\ \ref{fig:ablation_sens}\cvprcolor{C}, we analyze the impact of our proposed steerability loss $\mathcal{L}_\text{st}$ on the CB-SAE. We observe similar interpretability with and without $\mathcal{L}_\text{st}$, which is significantly better than the SAE baseline. And using $\mathcal{L}_\text{st}$ improves steerability by 2.9\%, validating its usefulness in CB-SAE training.

\begin{figure}[t]
    \centering
    \includegraphics[width=0.9\columnwidth]{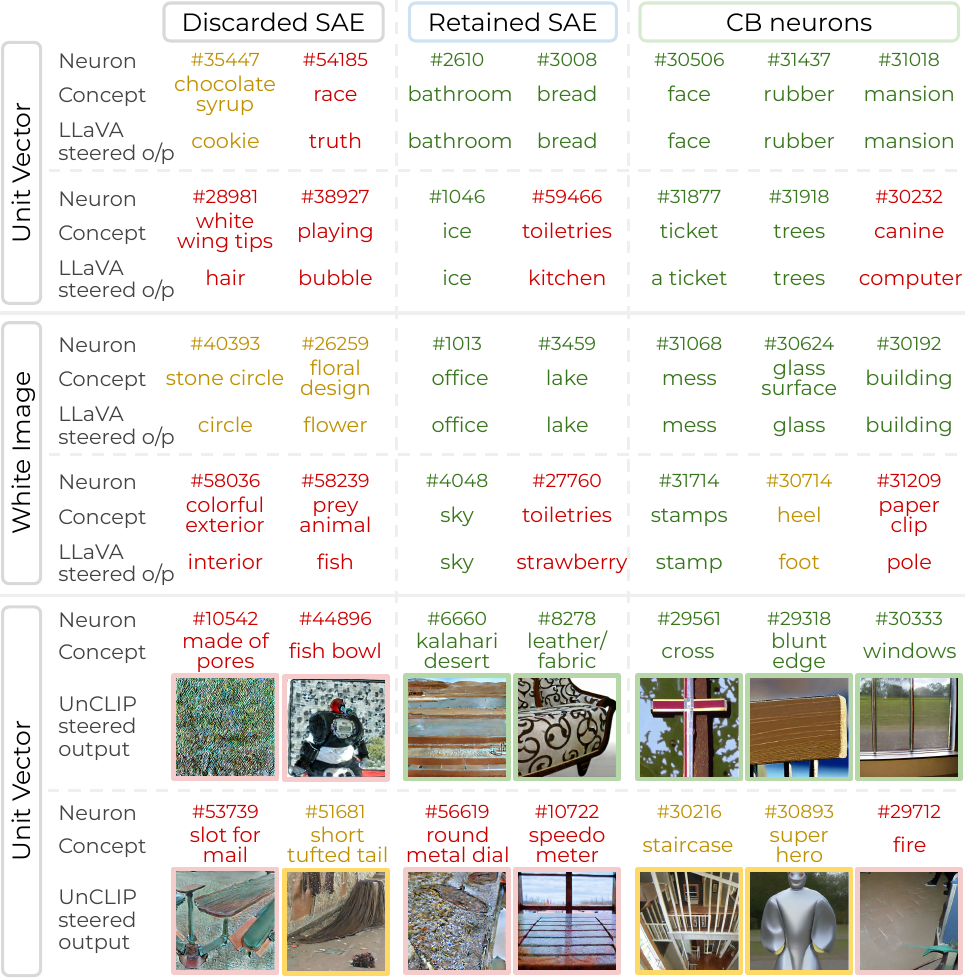}
    \vspace{-3mm}
    \caption{{Qualitative examples of steering UnCLIP and LLaVA. \darkgreen{Green} indicates successful steering, \textcolor[HTML]{bf9000}{yellow} indicates partial success, and \textcolor[HTML]{cc0000}{red} indicates failure cases. See Appendix for more results.}}
    \vspace{-3mm}
    \label{fig:qual_examples}
\end{figure}

\noindent
\textbf{Visualizing retained SAE and CB neurons.}
We visualize the distribution of CB neurons and retained SAE neurons \wrt interpretability and steerability scores in Fig.\ \ref{fig:cbsae_scatter}. Compared to Fig.\ \ref{fig:analysis}, the retained SAE neurons do not include low interpretability/steerability neurons as per our SAE pruning (Fig.\ \ref{fig:approach}, Step \cvprcolor{2}). Whereas the CB neurons have higher interpretability and similar steerability scores to the retained SAE neurons. Hence, there is potential to further improve steerability by designing better or more task-specific losses.

\noindent
\textbf{Qualitative examples of steering.}
We report qualitative examples of steering LLaVA and UnCLIP with discarded SAE neurons, retained SAE neurons, and CB neurons in Fig.\ \ref{fig:qual_examples}. We find discarded SAE neurons to be worse in steering than retained SAE neurons, which is expected as pruning uses the steerability score. Specifically, in the image-to-image generator UnCLIP \cite{Rombach_2022_CVPR}, we find steering the CB neurons produce higher quality images compared to the SAE neurons which often produce noisy images, likely due to the explicit concept supervision in CB-SAE. We also highlight some \textcolor[HTML]{cc0000}{failure cases} and \textcolor[HTML]{bf9000}{partially correct} steering for all neurons. 
\section{Conclusion}
In this work, we made the first attempt to unify two complementary paradigms - SAEs for unsupervised concept discovery and CBM for interpretable concept control - into a single unified framework, CB-SAE. Motivated by insights derived from our comprehensive analysis of SAEs in LVLMs, 
we first pruned the low-utility neurons in the SAE. 
We then introduced a light-weight CB module trained alongside the frozen, retained SAE using three principled objectives. 
Across vision-language (LLaVA) and image generation (UnCLIP) tasks, CB-SAE consistently improves interpretability and steerability, while enabling user-specified concept control.

\section*{Acknowledgements}
This work was performed under the auspices of the U.S.\ Department of Energy by the Lawrence Livermore National Laboratory under Contract No.\ DE-AC52-07NA27344, Lawrence Livermore National Security, LLC.\ and was supported by the LLNL-LDRD Program under Project No.\ 25-SI-001 and DOE ECRP 51917/SCW1885. LLNL-CONF-2013863. A.\ Kulkarni and T-W.\ Weng are also partially supported by National Science Foundation under Grant No.\ 2313105, 2430539, Hellman Fellowship, ARL Award, and Intel Rising Star Faculty Award.

\setcounter{figure}{7}
\setcounter{table}{2}
\appendix
\section*{Appendix}

In this appendix, we present full implementation details along with additional analyses. To support reproducibility, we will also release our codebase and pretrained models. The appendix is organized as follows:

\renewcommand{\labelitemii}{$\circ$}

\begin{itemize}
\setlength{\itemindent}{-0mm}
    \item Section~\ref{supp:sec:limitations}: Limitations and Future Work
    \item Section~\ref{sup:sec:implementation}: Implementation Details
    \begin{itemize}
        \setlength{\itemindent}{-0mm}
        \item Interpretability score
        \item CLIP-Dissect
        \item Cosine-cubed similarity loss
    \end{itemize}
    \item Section~\ref{sup:sec:expts}: Experiments
    \begin{itemize}
        \setlength{\itemindent}{-0mm}
        \item Experimental setup (Sec.\ \ref{sup:subsec:exptsetup}) 
        \item Interpretability vs steerability (Sec.\ \ref{sup:subsec:intvsst}, Fig.\ \ref{sup:fig:extanalysis})
        \item Extended analysis (Sec.\ \ref{sup:subsec:ext_analysis}, Table \ref{tab:sae_type_sensitivity}, \ref{tab:cbae_ablation}, \ref{sup:tab:pruning_score_sensitivity}, \ref{sup:tab:int_score_clip_sens}, Fig.\ \ref{sup:fig:cbtopk_sens})
        \item Extended qualitative results (Sec.\ \ref{sup:subsec:ext_qual_results}, Fig.\ \ref{sup:fig:ext_qual_examples})
    \end{itemize}
\end{itemize}{}

\section{Limitations and Future Work}
\label{supp:sec:limitations}
We acknowledge that the efficacy of our approach depends on the reliability of CLIP-Dissect in assigning accurate neuron-level concepts; however, continued advances in vision-language models are likely to enhance its performance. Extending and exploring hybrid approaches that combine the strengths of other unsupervised concept discovery methods such as transcoders~\cite{paulo2025transcoders} with user-specified concept control methods constitute our future work. 

\revision{Recent work on LLM SAEs \cite{chanin2025a} showed ``feature-splitting'' where hierarchical features split into fine-grained features (\eg ``math'' may be represented in the SAE by ``algebra'' and ``geometry'' neurons). It will be interesting to investigate if this phenomenon is connected to SAEs having limited concept coverage that we highlighted in this paper.}

\section{Implementation Details}
\label{sup:sec:implementation}


\noindent
\textbf{CLIP-Dissect \cite{oikarinen2023clip}.}
Consider a probing dataset of $N$ images $\mathcal{D}=\{x_i \in \mathbb{X}\}_{i=1}^N$ where $\mathbb{X}$ is the space of images, a concept set $\mathcal{C}=\{c_k\}_{k=1}^M$ with $M$ concepts in text form, and let layer $l$ of model $f$ being explained be denoted by $f_l$. CLIP-Dissect uses the probing set and a multimodal model, \eg CLIP \cite{radford2021learning} with an image and text encoder $E_I, E_T$ to identify concepts from $\mathcal{C}$ for individual neurons at the output of $f_l$. 

The probing set $\mathcal{D}$ is passed through the CLIP image encoder $E_I$ to obtain corresponding set of image embeddings $\{A_i = E_I(x_i)\}_{i=1}^N$. The concept set is passed through the CLIP text encoder $E_T$ to obtain text embeddings $\{E_T(c_k)\}_{k=1}^M$. Next, a matrix $P\in\mathbb{R}^{N\times M}$ is computed as the inner product of the image-text embeddings with entries $P_{ik}=A_i^\top E_T(c_k)$, as CLIP image and text encoders have the same embedding dimensions. The layer $l$ activations of a neuron $j$ for the same probing set are denoted by $q_j = [f_l(x_1)_j, f_l(x_2)_j, \cdots, f_l(x_N)_j]$. Finally, each neuron $j$ can be identified to have the concept $\arg\max_{k} \text{sim}(P_{:,k}, q_j)$ where $P_{:,k}$ is the $k^\text{th}$ column of $P$.
In other words, we compare each neuron's activations over the probing set with the corresponding activations of the CLIP model for each concept, and select the concept with the highest similarity. The maximum similarity itself (averaged across all neurons) is used as our \textit{interpretability score}.
The similarity function $\text{sim}$ is soft weighted pointwise mutual information (soft-WPMI) following \cite{oikarinen2023clip}. Please refer to the original paper \cite{oikarinen2023clip} for more details.

\noindent
\textbf{Cosine-cubed similarity loss \cite{oikarinen2023labelfree} $\mathcal{L}_\text{int}$.}
As discussed in Sec.\ \cvprcolor{5.2} (main paper), we use a cosine-cubed similarity loss $\mathcal{L}_\text{int}$ to train the CB encoder $E_\cb$ to produce concept predictions $c$ that match with CLIP zero-shot classifier predictions $\hat{y}$ for the same concept set $\mathcal{C}$. Concretely,
\begin{align}
    \mathcal{L}_\text{int}(c, \hat{y}) = \sum_{k=1}^{|\mathcal{C}|} -\frac{c_k^3\cdot \hat{y}_k^3}{\Vert c_k^3\Vert_2 \Vert \hat{y}_k^3 \Vert_2}
\end{align}
Here, $c_k$ is the $k^\text{th}$ concept prediction for the current mini-batch and $\hat{y}_k$ is the zero-shot CLIP prediction for concept $k$ with the same mini-batch. Following \cite{oikarinen2023labelfree}, we also normalize both vectors $c_k, \hat{y}_k\;\forall\; k$ before raising them to the third power (element-wise) and computing the cosine similarity. The third power is used to make the loss more sensitive to highly activating inputs. And we minimize the negative similarity which is equivalent to maximizing the similarity.

\begin{figure*}
    \centering
    \includegraphics[width=\linewidth]{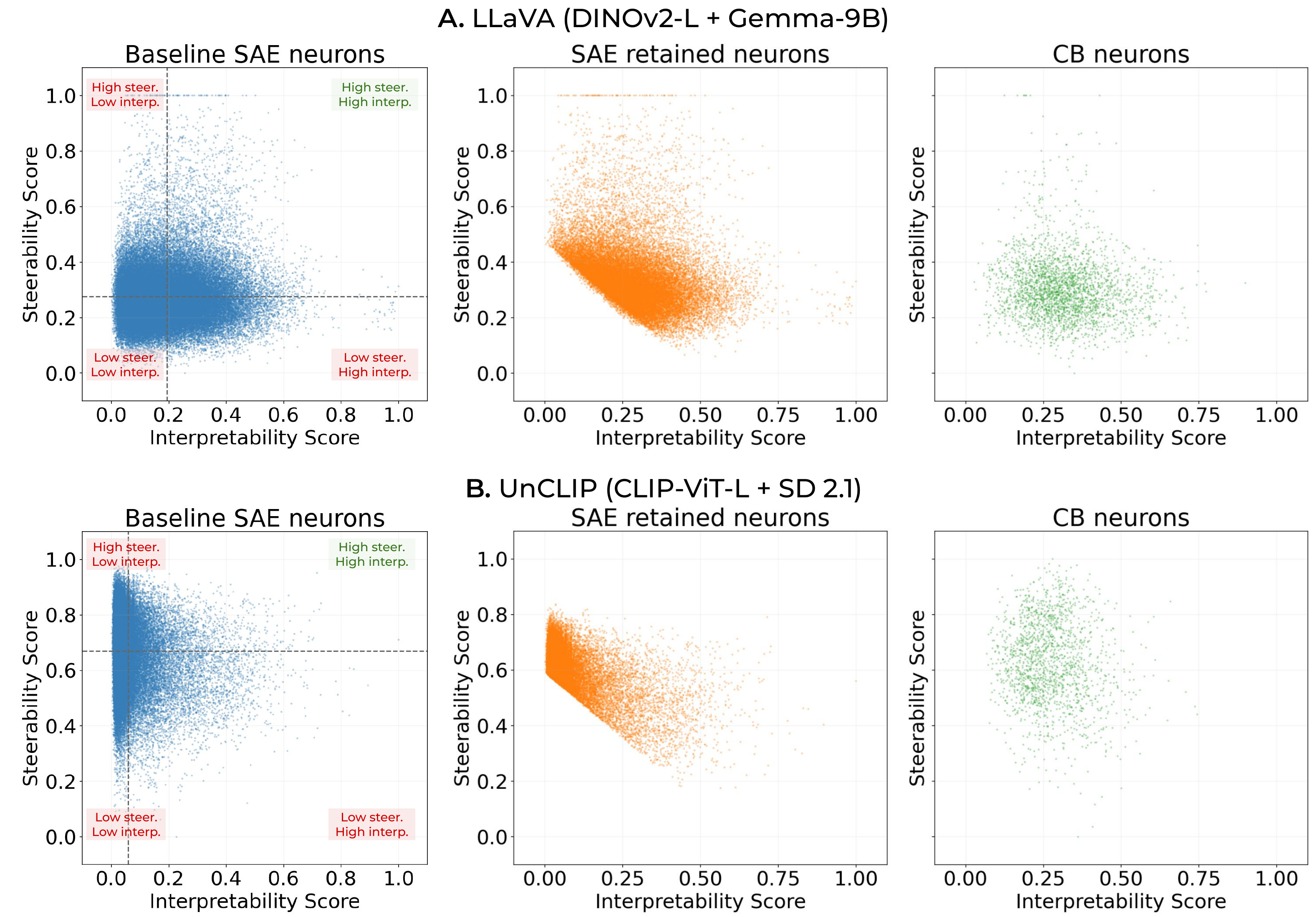}
    \caption{We analyze the interpretability and steerability of SAE and CB-SAE neurons for LLaVA with DINOv2 and Gemma2 as well as for UnCLIP with CLIP-ViT-L and Stable Diffusion 2.1. The dashed lines in the baseline SAE plots indicate the average scores along each axis.}
    \label{sup:fig:extanalysis}
\end{figure*}

\section{Experiments}
\label{sup:sec:expts}

\subsection{Experimental Setup}
\label{sup:subsec:exptsetup}

\textbf{Downstream model details.}
We experiment with SAEs/CB-SAEs trained on vision encoders for downstream models like LLaVA \cite{liu2023visual} and UnCLIP \cite{Rombach_2022_CVPR}. 
LLaVA models are large vision-language models that take an image and a text prompt as input and output a text-based answer (Fig.\ \cvprcolor{2A}, main paper).
Specifically, we used LLaVA-1.5-7B \cite{liu2024improved} which uses a CLIP-ViT-L-14-336 \cite{radford2021learning} vision encoder, a 2-layer MLP projector (not shown in Fig.\ \cvprcolor{2A} for simplicity), and an instruction-finetuned Vicuna-7B LLM \cite{vicuna2023}. We also use LLaVA-MORE \cite{cocchi2025llava} with DINOv2-Large \cite{oquab2024dinov2} vision encoder, a 2-layer MLP projector, and an instruction-finetuned Gemma2-9B LLM \cite{team2024gemma}.
On the other hand, UnCLIP is an image-to-image generative model that uses a CLIP-ViT-L \cite{radford2021learning} vision encoder and a finetuned Stable Diffusion 2.1 \cite{Rombach_2022_CVPR} as the image generator (Fig.\ \cvprcolor{2B}, main paper).

\noindent
\textbf{Miscellaneous details.}
We implement our CB-SAE in PyTorch \cite{paszke2019pytorch} building on the SAE codebase from \citet{pach2025sparse}. Following the baseline SAE training \cite{pach2025sparse}, we train the CB-SAE for 110k iterations with batch size 4096 and learning rate 2e$-$4 on a single 80GB Nvidia H100 GPU.

\begin{table*}[]
\centering
\caption{Sensitivity to type of SAE.}
\vspace{-3mm}
\label{tab:sae_type_sensitivity}
\begin{tabular}{lccccc}
\toprule
\multirow{2}{*}{SAE type} & \multicolumn{3}{c}{Interpretability} & \multicolumn{2}{c}{Steerability} \\
\cmidrule(lr){2-4}\cmidrule(lr){5-6}
 & CD & MS & Dead Neurons & Unit-Vec & White Image \\
 \midrule
Top-$k$ SAE & \textbf{0.162} & \textbf{0.548} & 52965 / 65536 & \textbf{0.228} & \textbf{0.241} \\
Batch Top-$k$ SAE & 0.158 & 0.540 & 56899 / 65536 & 0.226 & 0.231 \\
Matryoshka SAE & 0.154 & 0.517 & \textbf{4 / 65536} & 0.198 & 0.203 \\
\midrule
Top-$k$ CB-SAE & 0.264 & 0.556 & \textbf{0 / 32167} & \textbf{0.315} & \textbf{0.317} \\
Batch Top-$k$ CB-SAE & \textbf{0.265} & \textbf{0.564} & 0 / 32162 & 0.307 & 0.299 \\
Matryoshka CB-SAE & 0.244 & 0.556 & 4 / 32169 & 0.261 & 0.250 \\
\bottomrule
\end{tabular}
\end{table*}

\begin{table}[]
\centering
\caption{Ablation to quantify the usefulness of SAE in CB-SAE.}
\vspace{-3mm}
\label{tab:cbae_ablation}
\resizebox{\linewidth}{!}{
\begin{tabular}{lccc}
\toprule
\multirow{2}{*}{} & Interpretability & \multicolumn{2}{c}{Steerability} \\
\cmidrule(lr){2-2}\cmidrule(lr){3-4}
 & CLIP-Dissect score & Unit-Vec & White Image \\
\midrule
SAE \cite{pach2025sparse} & 0.154 & 0.198 & 0.203 \\
CB-AE (w/o SAE) & \textbf{0.308} & 0.238 & 0.232 \\
CB-SAE (Ours) & 0.244 & \textbf{0.261} & \textbf{0.250} \\
\bottomrule
\end{tabular}
}
\end{table}

\subsection{Interpretability vs Steerability in SAEs}
\label{sup:subsec:intvsst}

We extend our analysis from Sec.\ \cvprcolor{4} (main paper) on an SAE from LLaVA with CLIP image encoder to SAEs from LLaVA with DINOv2 image encoder and UnCLIP image-to-image generation model with CLIP image encoder in Fig.\ \ref{sup:fig:extanalysis} (left). We report our observations (repeating those from Sec.\ \cvprcolor{4}):
\begin{itemize}
    \itemsep0em
    \item LLaVA (CLIP-ViT-L + Vicuna-7B, Fig.\ \cvprcolor{3}, main paper):
    \begin{itemize}
        \itemsep0em
        \item Low interpretability, low steerability: 36.26\% (23763)
        \item High interpretability, low steerability: 19.87\% (13022)
        \item Low interpretability, high steerability: 25.03\% (16403)
        \item High interpretability, high steerability: 18.84\% (12348)
        \end{itemize}
    \item LLaVA (DINOv2-L + Gemma-9B, Fig.\ \ref{sup:fig:extanalysis}\cvprcolor{A}):
    \begin{itemize}
        \itemsep0em
        \item Low interpretability, low steerability: 33.07\% (21675)
        \item High interpretability, low steerability: 23.35\% (15304)
        \item Low interpretability, high steerability: 23.75\% (15565)
        \item High interpretability, high steerability: 19.82\% (12992)
        \end{itemize}
    \item UnCLIP (CLIP-ViT-L + Stable Diffusion 2.1, Fig.\ \ref{sup:fig:extanalysis}\cvprcolor{B}):
    \begin{itemize}
        \itemsep0em
        \item Low interpretability, low steerability: 30.84\% (20209)
        \item High interpretability, low steerability: 14.53\% (9517)
        \item Low interpretability, high steerability: 42.76\% (28022)
        \item High interpretability, high steerability: 11.88\% (7788)
        \end{itemize}
\end{itemize}
Note that the average steerability score for UnCLIP is higher than for LLaVA since the scores are computed in image embedding space and text embedding space respectively. 
Across both types of models, we consistently find that only a small portion of neurons (12-20\%) are useful for both interpretability and steerability. And a majority of neurons (30-36\%) are unsuitable for both interpreting new inputs and 
steering outputs.

We also show the retained SAE neurons and CB neurons in Fig.\ \ref{sup:fig:extanalysis} (right) similar to Fig.\ \cvprcolor{6} (main paper). We find CB neurons are similar to retained SAE neurons while being significantly better than the discarded SAE neurons (also shown quantitatively in Table \cvprcolor{1}, \cvprcolor{2}, main paper). We emphasize that CB neurons have to incorporate relatively more difficult concepts due to our concept set selection (Sec.\ \cvprcolor{5.2}, main paper) which excludes already discovered (and relatively easier to learn) concepts present in the retained SAE. Hence, it is more difficult for CB neurons to always outperform the retained SAE neurons.

\begin{table*}[]
\centering
\caption{\textbf{Sensitivity to choice of metrics for SAE pruning.}}
\vspace{-3mm}
\label{sup:tab:pruning_score_sensitivity}
\resizebox{0.9\linewidth}{!}{
\begin{tabular}{@{}lccccc@{}}
\toprule
\multirow{2}{*}{Scores for SAE pruning} & Reconstruction evaluation & \multicolumn{2}{c}{Interpretability evaluation} & \multicolumn{2}{c}{Steerability evaluation} \\ \cmidrule(lr){2-2}\cmidrule(lr){3-4}\cmidrule(lr){5-6} 
 & Zero-shot ImageNet Acc. (\%) & CLIP-Dissect & Monosemanticity & Unit Vector & White Image \\ \midrule
None (SAE baseline) \cite{pach2025sparse} & \textbf{74.07} & 0.154 & 0.517 & 0.198 & 0.203 \\
Interpretability score only & 73.39 & \underline{0.233} & \textbf{0.566} & 0.216 & 0.220 \\
Steerability score only & 70.99 & 0.167 & 0.520 & \textbf{0.288} & \textbf{0.269} \\
Both scores & \underline{73.78} & \textbf{0.244} & \underline{0.556} & \underline{0.261} & \underline{0.250} \\ \bottomrule
\end{tabular}
}
\end{table*}

\begin{table}[t]
\centering
\caption{Sensitivity of interpretability evaluation with CLIP-Dissect to choice of CLIP-like model used.}
\vspace{-3mm}
\label{sup:tab:int_score_clip_sens}
\resizebox{\linewidth}{!}{
\begin{tabular}{@{}llcc@{}}
\toprule
\multicolumn{2}{l}{CLIP-like model for evaluation} & \multicolumn{2}{c}{Interpretability Score} \\ \cmidrule(lr){1-2} \cmidrule(lr){3-4}
Model & Architecture & SAE & CB-SAE (\textit{Ours}) \\
\midrule
CLIP \cite{radford2021learning} & ViT-B-16 & 0.198 & \textbf{0.307} \\
CLIP \cite{radford2021learning} & ViT-L-14-336 & 0.154 & \textbf{0.244} \\
SigLIP \cite{zhai2023sigmoid} & ViT-SO400M-14-384 & 0.189 & \textbf{0.289} \\
SigLIP2 \cite{tschannen2025siglip} & ViT-gopt-16-384 & 0.188 & \textbf{0.290} \\
SigLIP2 \cite{tschannen2025siglip} & ViT-SO400M-16-384 & 0.176 & \textbf{0.272} \\
DFN \cite{fang2024data} & ViT-H-14-378 & 0.220 & \textbf{0.347} \\
PE-core \cite{bolya2025perception} & BigG-14-448 & 0.207 & \textbf{0.312} \\ \bottomrule
\end{tabular}
}
\end{table}

\noindent
\revision{\textbf{Analysis using null baseline for steering scores.}
In the previous analysis of interpretability vs steerability in SAEs, we used the average (or mean) score as the threshold for both metrics. Another choice could be a null baseline for steerability, which would be the average steerability score of the original model's neurons without an SAE. We found this steerability score of CLIP layer 22 neurons (which are the input to LLaVA SAE/CB-SAE) to be 0.173, while the average steerability score in Fig.\ \cvprcolor{3} (main paper) was 0.232. Using 0.173 as the steerability threshold, the proportions of neurons change as follows:
\begin{itemize}
    \itemsep0em
    \item LLaVA (CLIP-ViT-L + Vicuna-7B, Fig.\ \cvprcolor{3}, main paper):
    \begin{itemize}
        \itemsep0em
        \item Low interp., low steer.: 36.26\% $\to$ 17.86\%
        \item High interp., low steer.: 19.87\% $\to$ 9.58\%
        \item Low interp., high steer.: 25.03\% $\to$ 43.42\%
        \item High interp., high steer.: 18.84\% $\to$ 29.14\%
        \end{itemize}
\end{itemize}
However, our original claim of SAEs having low proportion (29\%) of high utility neurons is still valid.
}

\noindent
\revision{\textbf{Traditional SAE metrics for 4 neuron classes.}
Reconstruction loss $||v-\hat{v}||_2^2$ (Sec.\ \cvprcolor{3}, main paper) does not directly involve SAE latent neurons. To isolate and quantify their impact, we individually zero each SAE neuron and compute the reconstruction loss change \wrt the original SAE. Then, the average loss changes over Interpretability/Steerability neuron classes (from Fig.\ \cvprcolor{3}) are: low/low (1.6e-5), high/low (1.4e-6), low/high (1.1e-5), high/high (3.9e-6). 
We observe low interpretability neurons contribute more to reconstruction, while steerability is relatively less influential.
}

\subsection{Extended Analysis of our CB-SAE}
\label{sup:subsec:ext_analysis}

\noindent
\revision{\textbf{Sensitivity to type of SAE.} In Table \ref{tab:sae_type_sensitivity}, we evaluate the sensitivity of our CB-SAE to the type of pretrained SAE used. We consider Top-$k$ and Batch Top-$k$ SAEs in addition to the Matryoshka SAEs already compared in the main paper. We find that our CB-SAE can provide consistent improvements regardless of the type of pretrained SAE used. Interestingly, Top-$k$ and Batch Top-$k$ SAEs have better interpretability and steerability than Matryoshka SAEs, but also feature a very high number of dead neurons, \ie SAE neurons which do not activate for any inputs. This makes sense since Matryoshka SAEs were proposed to overcome the dead neurons limitation. Further, our CB-SAE can also resolve the dead neurons problem by eliminating frequently activating SAE neurons.}

\noindent
\revision{\textbf{Ablation of SAE from CB-SAE.} In Table \ref{tab:cbae_ablation}, we evaluate a CB-SAE model without using any SAE, \ie a CB-AE \cite{kulkarni2025interpretable} where all the user-defined concepts are directly used in the concept bottleneck. We find that CB-AE has higher interpretability score than CB-SAE since all concepts are now explicitly optimized for it. On the other hand, CB-SAE achieves better steerability since the retained SAE neurons have high steerability based on our analysis.}

\noindent
\textbf{Sensitivity to scores used for SAE pruning.} We extend our sensitivity analysis from Fig.\ \cvprcolor{5A} (main paper) in Table \ref{sup:tab:pruning_score_sensitivity} to additionally include monosemanticity score \cite{pach2025sparse} (interpretability evaluation) and zero-shot ImageNet-1k accuracy (reconstruction evaluation) when using the SAE/CB-SAE reconstructed latents. We observe that using either the interpretability score or both scores yields similar reconstruction as the baseline SAE, while steerability-based pruning leads to significantly worse reconstruction. Similarly, using either the interpretability score or both scores improves the monosemanticity significantly \wrt the baseline, while steerability-based pruning provides only a marginal gain over the baseline.

\noindent
\textbf{Sensitivity to CLIP model in interpretability evaluation.} We evaluate the sensitivity of our interpretability evaluation with CLIP-Dissect by varying the CLIP-like model used, in Table \ref{sup:tab:int_score_clip_sens}. While our evaluation used a stronger CLIP-ViT-L-14-336 \cite{radford2021learning} model \wrt the smaller CLIP-ViT-B-16 used for training the CB-SAE, we now evaluate with even stronger models including SigLIP \cite{zhai2023sigmoid}, SigLIP2 \cite{tschannen2025siglip}, Data Filtering Networks (DFN) \cite{fang2024data} and Perception Encoder (PE) \cite{bolya2025perception}. Across all CLIP-like models, our CB-SAE achieves consistent gains over the baseline SAE for LLaVA with CLIP-ViT-L encoder, validating that our choice of CLIP-like model for interpretability score does not affect our evaluation.

\begin{figure}[t]
    \centering
    \includegraphics[width=\linewidth]{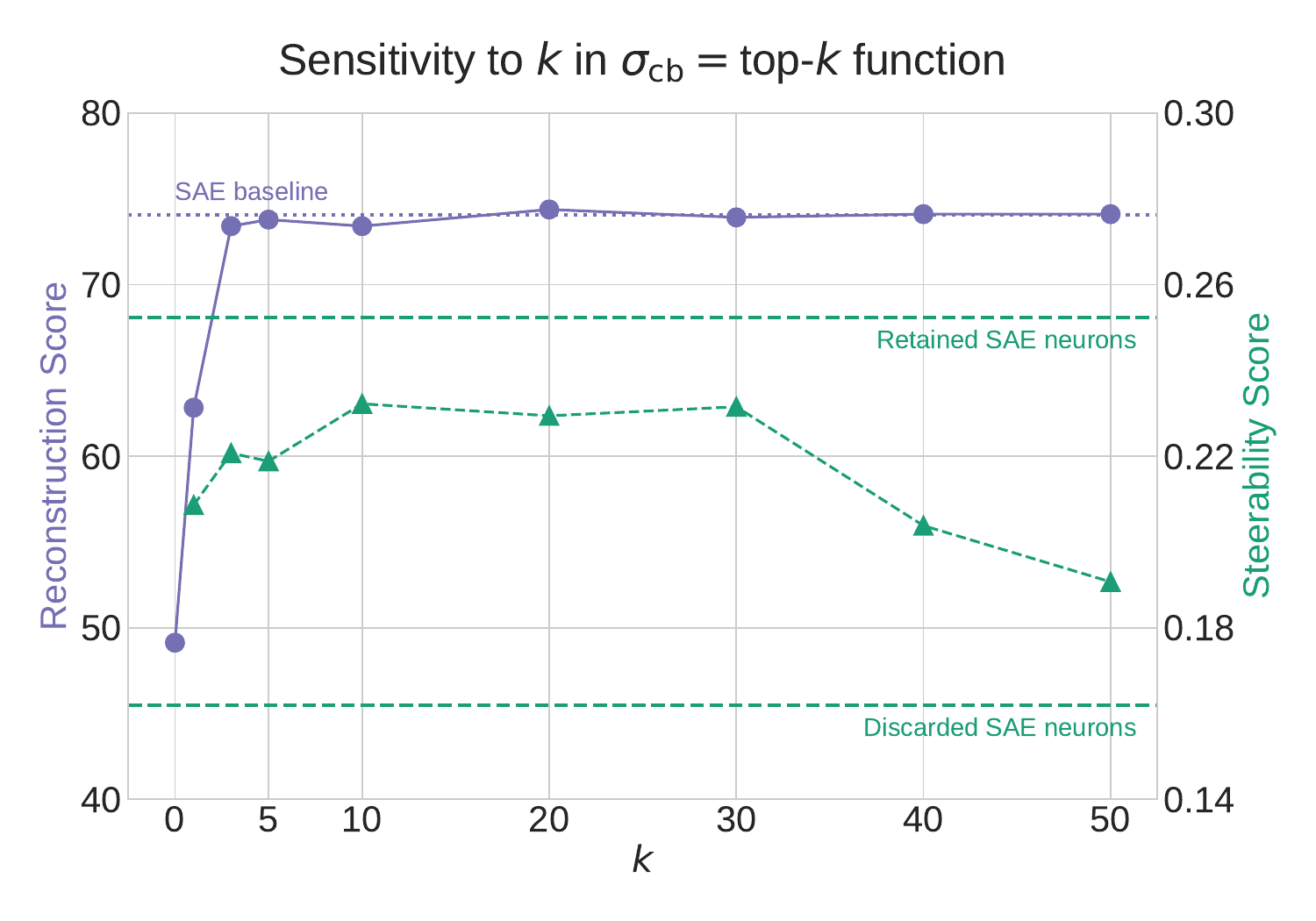}
    \vspace{-6mm}
    \caption{Sensitivity analysis of CB-SAE in LLaVA to $k$ in top-$k$ activation function used in the CB decoder. Steerability score here is computed only for CB neurons, reconstruction score is zero-shot accuracy when using SAE/CB-SAE reconstructions of CLIP latents on ImageNet-1k.}
    \label{sup:fig:cbtopk_sens}
\end{figure}

\noindent
\textbf{Sensitivity to $k$ in $\sigma_\cb$.} In Fig.\ \ref{sup:fig:cbtopk_sens}, we analyze the sensitivity of our CB-SAE to the choice of $k$ in the top-$k$ activation function used in the CB decoder. Here, we define reconstruction score as the zero-shot ImageNet-1k accuracy of CLIP when using SAE/CB-SAE reconstructed latents. We also report the white image steerability score of only the CB neurons to understand the impact of $k$ on steerability. Note that we do not consider interpretability score here since $\sigma_\cb$ is only applied in the CB decoder while interpretability evaluation only considers the CB encoder, \ie interpretability score does not change when varying $k$. We observe that reconstruction score improves as $k$ increases, but it is already very close to the baseline even at $k=3$ to $k=5$. The steerability score first increases with $k$ and then decreases for $k>30$. This is because with higher $k$, steering might be less successful as the selected concept contends with many other concepts to be combined into the final reconstructed latent. On the other hand, if $k$ is too low, then the reconstruction might not be good enough for the downstream model to produce the appropriate response. However, across all values of $k$, our CB-SAE is able to outperform the discarded SAE neurons while being worse than the retained SAE neurons. Hence, future work can develop more steerability-focused training objectives to further improve steerability.

\noindent
\revision{\textbf{Correlation between LLaVA and UnCLIP steerability.}
We used the same training method for CB-SAEs in both LLaVA and UnCLIP, but existing pretrained LLaVA and UnCLIP models do not use the same CLIP model (and even use different layers of CLIP and different number of tokens). However, we computed the Pearson correlation between the steerability scores of 1329 common CB concepts from LLaVA and UnCLIP CB-SAEs, which is 0.06 with a p-value 0.035. Hence, steerability seems to be largely downstream task/model-dependent.
}

\begin{figure*}
    \centering
    \includegraphics[width=0.8\linewidth]{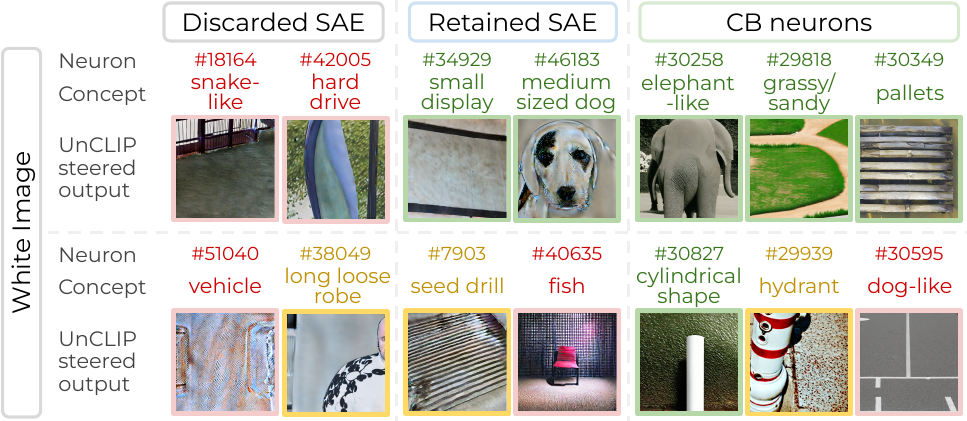}
    \caption{Qualitative examples of steering UnCLIP. \darkgreen{Green} indicates successful steering, \textcolor[HTML]{bf9000}{yellow} indicates partial success, and \textcolor[HTML]{cc0000}{red} indicates failure cases.}
    \label{sup:fig:ext_qual_examples}
\end{figure*}

\subsection{Extended Qualitative Results}
\label{sup:subsec:ext_qual_results}

We provide qualitative examples of white image steering of UnCLIP with SAE/CB-SAE in Fig.\ \ref{sup:fig:ext_qual_examples}. Similar to our results in Fig.\ \cvprcolor{7} (main paper), we find steering CB-SAE neurons produces higher quality images while SAE neurons tend to produce more noisy images.

{
    \small
    \bibliographystyle{ieeenat_fullname}
    \bibliography{main}
}

\end{document}